# Colour and Brush Stroke Pattern Recognition in Abstract Art using Modified Deep Convolutional Generative Adversarial Networks

Srinitish Srinivasan, Varenya Pathak, Abirami S

School of Computer Science and Engineering, Vellore Institute of Technology, Chennai

# Abstract

Abstract Art is an immensely popular, discussed form of art that often has the ability to depict the emotions of an artist. Many researchers have made attempts to study abstract art in the form of edge detection, brush stroke and emotion recognition algorithms using machine and deep learning. This papers describes the study of a wide distribution of abstract paintings using Generative Adversarial Neural Networks(GAN). GANs have the ability to learn and reproduce a distribution enabling researchers and scientists to effectively explore and study the generated image space. However, the challenge lies in developing an efficient GAN architecture that overcomes common training pitfalls. This paper addresses this challenge by introducing a modified-DCGAN (mDCGAN) specifically designed for high-quality artwork generation. The approach involves a thorough exploration of the modifications made, delving into the intricate workings of DCGANs, optimisation techniques, and regularisation methods aimed at improving stability and realism in art generation enabling effective study of generated patterns. The proposed mDCGAN incorporates meticulous adjustments in layer configurations and architectural choices, offering tailored solutions to the unique demands of art generation while effectively combating issues like mode collapse and gradient vanishing. Further this paper explores the generated latent space by performing random walks to understand vector relationships between brush strokes and colours in the abstract art space and a statistical analysis of unstable outputs after a certain period of GAN training and compare its significant difference. These findings validate the effectiveness of the proposed approach, emphasising its potential to revolutionise the field of digital art generation and digital art ecosystem.

# I. Introduction

Pattern recognition is based on the generalisation of objects to classify real-world items by identifying certain shared common features. Pattern recognition, a machine learning algorithm, considers useful features while eliminating redundant ones [1]. This technique is invaluable in various fields such as text pattern recognition [2], fingerprint scanning [3], seismic activity analysis [4], audio recognition [5] and healthcare [6].

In our study, pattern recognition was implemented through Generative Modelling [7] using Generative Adversarial Networks (GANs) [8]. Generative Models have the ability to learn and explain the distribution large amount of data in various forms such as audio, images, words, etc. Once the model is trained, it can generate new data by extracting samples from the derived data or from random noise. These models have to ability to learn and model underlying categories, dimensions, and other aspects without specific programming. There are different generative models tailored for specific requirements, including GANs for style transfer in images [9], Hidden Markov Models (HMMs) for speech recognition [10], Variational Auto-encoders (VAEs) for image generation [11], and Auto-encoders for anomaly detection [12].

Generative Adversarial Networks(GANs) consist of a generator and a discriminator. The generator produces images after learning the distribution while the discriminator, a Deep Neural Network [13],determines whether the generated image is real or fake. GANs are chosen for their ability to produce high-quality outputs from grainy inputs and for their flexibility and fine-tuning capabilities, crucial aspects of any model.

This study explores and derives the mathematical features of colour patterns through latent space exploration by random walks [14]. A 1D random walk involves an object moving left, right, or staying in place, with probabilities determining its future movements. In a 2D random walk, the movement extends to the XY, YZ, or XZ axis—up, down, left, or right—with equal probabilities, akin to the movement of chess pieces on a board. In a 3D random walk, spatial positions are considered, predicting the object's probability in a Monte Carlo [15] randomised algorithm.

Random walks are implemented in this study to support style transfer and fusion, where a random walk in a latent space mixes various styles smoothly. This approach starts from a known point in the latent space [16] and explores random directions, generating unique art styles while mapping artistic parameters provided by the user, such as textures and brush stroke styles. These mapped parameters serve as inputs for the model, guiding it through random directions in the latent space. This method also generates a sequence of latent vectors, enabling a gradual transformation and evolution of the generated images. The transitions between styles in our outputs are gradual and exploratory providing state of the art results enabling us to successfully study abstract art patterns. Further this paper includes the study of distorted patterns after a certain point in training and describes the statistical analysis and tests performed in order to compare the distorted colour space with the original space.

The following sections of the paper are divided as follows: **Section II** describes the related work with respect to pattern recognition and Generative Adversarial Networks(GAN). **Section III** describes the proposed workflow and architecture of our study. **Section IV** discusses the results of the brush stroke colour patterns qualitatively, latent space exploration by random walks and a statistical analysis of distorted outputs at a certain period of training. **Section V** concludes our study and describes the future scope of this study.

# II. Related Work

In this section, we discuss related literature with respect to pattern recognition using Machine learning techniques, and the use of GANs of in several pattern recognition, generative and style transfer tasks.

[17] studied the spatial and temporal variations of water quality of the Suquia River Basin. The researchers used factor analysis, cluster analysis, principal component analysis and Discriminant Analysis to cluster and obtain patterns. They concluded that Cluster Analysis gives good results as an initial exploratory method to evaluate spatial and temporal differences and reduces the number of parameters by 77% to differentiate between four spatial areas and Discriminant Analysis provides the best results for temporal and spatial analysis. It performs 73% reduction in parameters to differentiate between wet and dry season upto a precision of 87%. [18] studied different pattern recognition techniques for apple sorting. They made use of K-Nearest Neighbours algorithm using 1 and 2 nearest neighbours , decision trees  and Artificial neural networks. They noted that with respect to classification performance, 1 and 2 nearest neighbours methods using five input features yielded the second best results while the Neural Network was able to detect non linear relationships in apple sorting patterns.

In the field of pattern recognition and classification with respect to art works, [19] studied various pattern extraction techniques using Generative Adversarial Neural Networks and Deep Convolutional Neural Networks to classify art periods, emotions from art works and construct a social network of artists. [20] used a Radial Basis Function neural network classifier to model western paintings for classification. These groups of networks are very powerful and have been used in function approximation, pattern classification and data compression. For the feature extraction process, the researchers made use of Gabor wavelets, a popular wavelet transform in image processing [21]. [22] performed a 3-step hierarchal classification of paintings using face and brush stroke models. Their 3-step approach is inclusive of colour classification by grouping portrait miniatures by computing the mean RGB value, followed by shape classification on a region by region basis by reducing the search space to a specific Region of Interest followed by stroke detection and classification. [23] studied fast texture synthesis using tree-structured vector quantisation. They implemented a Gaussian Pyramid and Markov Random Field like architecture and used tree-structured vector quantisation for acceleration, a common method for data compression. The approach has the ability to replicate an image on given texture as input. [24] studied style and abstraction in portrait sketching. The researchers replicate the sketch stroke of artists by performing edge detection using the canny edge detection operator in addition to stroke matching and curve detection. [25] analyses various algorithms and methods for stroke-based rendering. The optimisation method includes Voronoi Algorithms that use the property of SBR to perform efficient global update steps, trial and error algorithms performs heuristically chosen tests to reduce randomness. The researchers studied Greedy algorithms were studied as well but it was concluded that they are too slow for any interactive application. [26] proposed features derived from colours, edges and grey scale-texture of images that discriminate paintings from photographs. They proposed a neural-network classification methodology with 6 sigmoidal units in a unique hidden layer to perform painting-photograph discrimination.

Generative Adversarial Neural networks, since its introduction by [8] has become popular in the field of art being used for generation and style transfer purposes. [27] compared various popular GAN architectures.

They concluded that Pix2Pix could be relevant for contemporary simple-styled style transfer tasks for Ortho-images but not suitable for old map styles which are more different and visually complex in content and styles, while Cycle-GAN could be more revenant for such images. [28] uses a 5-layer CNN to perform style transfer on images. They noted that the speed of image synthesis is hindered by the image and resolution of the fearer space, in addition to this they mentioned that denoising images is a challenge with this architecture. [29] implemented UnityGAN to learn the style changes between camera, producing shape-stable style unity images for each camera. They made use of skip-connections between multi-depth layers which enabled the retention of more structural information therefore accoutring for the stability of the generated image. [30] proposed **APDrawingGAN++** to transform the photo of a face to a high quality APDrawing. It made use of auto encoders to improve facial feature drawings, lip and hair classifiers were introduced to guide the local generator and auto encoder to a desired style. Moreover the researchers made use of DT loss to penalise large misalignments. [31] compared and analysed various kinds of Neural networks for art-based applications such as GANS, Image stylisation, DeepDream and Perception Engines which includes image Fourier models. [32] implements BigGAN- deep model on the ImageNET dataset with hierarchical latent spaces. BigGAN deep differs from the BigGAN model as it contains 2 extra 1X1 convolutions to provide the required number of output filters for the images. On increasing the depth by two, the researchers noted performances were negatively affected to an extent. [33] used **mGANprior** that employs multiple latent codes for reconstructing real images with a pre trained-GAN model. It enables the use of GAN's as a powerful prior for pre-processing tasks such as colorisation, in painting, inverting images. [34] implements DCGAN and finds closest latent features in order to update the latent vector gradually and smoothly to generate the desired image. The architecture was used to make desired edits to images based on users requirements.[35] proposed an **InterfaceGAN** to interpret the semantics encoded in the latent space of GANs. Provides a rigorous analysis of the semantic attributes emerging in the latent space of well-trained GAN models, and then constructs a manipulation pipeline of for leveraging the semantics in the latent code for facial attribute editing. The architecture is tested against encoder-decoder generative models and StyleGANs. [36] implements a progressive GAN growth experiment for improved quality in GAN outputs. The researchers start from a low resolution, and progressively add new layers. The researchers tested the performance of the model on CIFAR10 with an inception score of 8.80.

# III. Proposed Work

## 1. Pre-Processing

The images used for the study have different sizes such as 1024X2048, 512X512, 2048X2048 etc. They were resized to a standard size of 256X256, the average dimensions of the images. In the resized images, standard noise filtering techniques such as Gaussian and Median filters are applied in order to filter out additive gaussian noise. The standard deviation used for the gaussian filter was 0.001, to ensure the process does not lose features through blurring. Further, normalisation is performed by calculating a z-score for all 3 channels. Equation 3.1 describes the expression used for normalisation.

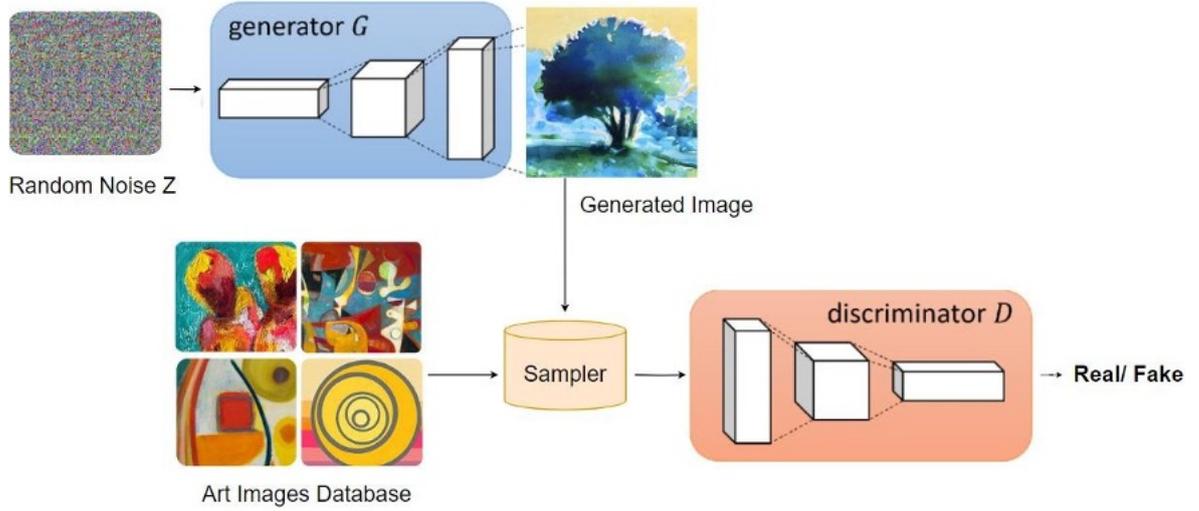

Figure 1: Workflow of the proposed model

$$z_i = \frac{I_{ij} - \mu_i}{\sigma_i} \quad (1)$$

In equation (3.1), $i$ represents the channel index which ranges from 1 to 3, $j$ represents the pixel which has a range from 0 to 255. The pixel values are subtracted from the mean of each channel and divided by the standard deviation. It is to note that no data augmentation was performed such as random rotation, vertical or horizontal flip, channel flipping etc due to presence of over 2500 images and similar colour patterns on the edges of each painting.

## 2. Architecture

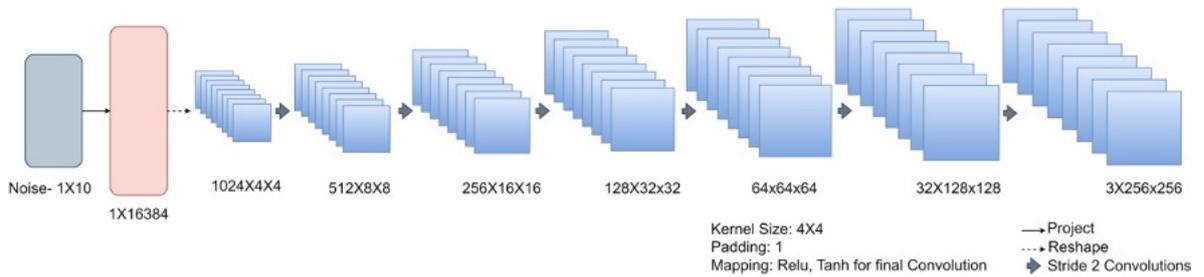

Figure 2(a): Generator Architecture

In this section, we discuss about the architecture of mDCGAN proposed for art image generation. The proposed work involves modifications to both the generator and discriminator components of a DC-GAN architecture, aimed at achieving stable and enhanced art image generation. A GAN consists of two deep neural network models, a generator and discriminator. The generator tries to overcome the discriminator by trying to make the generator predict all its outputs as real whereas the discriminator tries to distinguish between real and fake images, setting up an adversarial scenario as per game theory[37]. In this section, we discuss the architecture modifications done for this study. The modified generator and discriminator layers of the mDCGAN architecture is as shown in Figure 2(a) and 2(b).

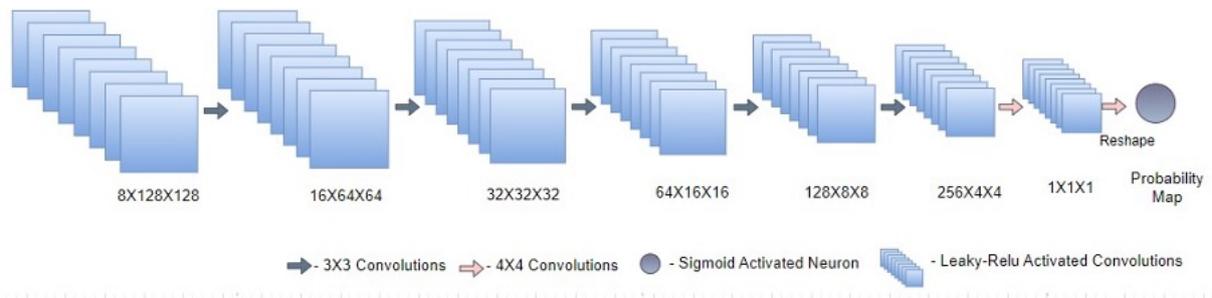

Figure 2(b): Discriminator Architecture

The architectural modifications of mDCGAN draw inspiration from [38], with adaptations tailored to accommodate 256x256 image dimensions. These architectural modifications of mDCGAN aim to enhance the stability and diversity of art image generation, accommodating the unique demands of generating art while mitigating common challenges associated with GANs such as modal collapse, and noisy outputs. A 4X4 kernel for the final convolution layer, a final reshape layer and, additional dropouts are added at the discriminator to ensure the discriminator does not overfit the generator. As suggested in DC-GAN, mDCGAN uses leaky ReLU activated functions with negative slope of 0.2 for the discriminator. The final layer activation of the generator was mapped using tanh(z) activation function, where z represents the output of a convolution mapping. In mDCGAN, square kernels of length 4 is used for every layer of the generator, with padding set to 1. A convolution layer in the generator consists of (i) Transpose convolution layers of stride 2 and padding 1, (ii) A Batch normalisation layer (iii) ReLU[39] activation function for all layers except the last layer which uses tanh. The number of kernels decrease by a factor of 2 for every layer to construct an image with 3 channels. The first 2 layers in the generator are linear and reshape[citation] layers respectively, aimed to transform a 100-dimensional vector to a vector with the help of a linear transformation. The 16384-feature vector is reshaped to a block of size 1024X4X4, where 1024 represents the number of kernel filters or channels and 4X4 represents the height and width dimensions of the embedding. After reshaping the embeddings are transpose convolved through 6 layers to construct an image of size 3X256X256 where 3 represents the RGB channels. For the discriminator, mDCGAN uses square kernels of size 3 for every layer with padding set to 1. A convolution layer in the discriminator consists of (i)Convolution mapping of stride 2 and padding 1, (ii) Batch normalisation layer (iii)Leaky ReLU activated functions with negative slope set as 0.2. For every layer, the number of kernels increase by a factor of 2 which enables the model to learn low level features at a smaller spatial dimension. The final convolution layer of the discriminator uses a square kernel of size 4, this plays a role in reducing the dimensions to 1X1X1 with fewer convolution layers, ensuring the discriminator does not overfit the generator due to a deeper network. The final layer of the discriminator is a reshape layer that transforms an image of size

1X1X1 into a one-dimensional vector, so that it can be passed into a probability mapping function such as sigmoid. Table 3.1 and 3.2 describe the layers of the discriminator and generator respectively.

## 3. Training Workflow

During training, labels are assigned for real and fake images. The images from the dataset are labelled as real and the images generated from noise is labelled as fake. The role of the discriminator is to accurately classify all real images as real and fake images as fake. First the output of the discriminator model is computed on an equal distribution of real and fake images. Then the loss is estimated to update the discriminator weights. Then a sample of 100 data points, represented as a 100-dimensional vector is obtained from a uniform random distribution and passed into the generator. The generated samples, output of the generator, are passed into the discriminator which maps them to real labels (real/fake). That is, the generator tries to convince the discriminator that its generated image is real. In this process, if the generator successfully does so, the generator is rewarded, else it's penalised. The loss value is computed and the weights for the generator are updated. This sets up an adversarial training loop.

Table 1(a): Discriminator Layers

| Layer | Output Size | Parameters |
|---|---|---|
| Conv2d-1 | [8, 8, 128, 128] | 224 |
| BatchNorm2d-2 | [8, 8, 128, 128] | 16 |
| Dropout2d-3 | [8, 8, 128, 128] | 0 |
| LeakyReLU-4 | [8, 8, 128, 128] | 0 |
| Conv2d-5 | [8, 16, 64, 64] | 1,168 |
| Dropout2d-6 | [8, 16, 64, 64] | 0 |
| BatchNorm2d-7 | [8, 16, 64, 64] | 32 |
| LeakyReLU-8 | [8, 16, 64, 64] | 0 |
| Conv2d-9 | [8, 32, 32, 32] | 4,640 |
| Dropout2d-10 | [8, 32, 32, 32] | 0 |
| BatchNorm2d-11 | [8, 32, 32, 32] | 64 |
| LeakyReLU-12 | [8, 32, 32, 32] | 0 |
| Conv2d-13 | [8, 64, 16, 16] | 18,496 |
| Dropout2d-14 | [8, 64, 16, 16] | 0 |
| BatchNorm2d-15 | [8, 64, 16, 16] | 128 |
| LeakyReLU-16 | [8, 64, 16, 16] | 0 |
| Conv2d-17 | [8, 128, 8, 8] | 73,856 |
| Dropout2d-18 | [8, 128, 8, 8] | 0 |
| BatchNorm2d-19 | [8, 128, 8, 8] | 256 |
| LeakyReLU-20 | [8, 128, 8, 8] | 0 |
| Conv2d-21 | [8, 256, 4, 4] | 2,95,168 |
| Dropout2d-22 | [8, 256, 4, 4] | 0 |
| BatchNorm2d-23 | [8, 256, 4, 4] | 512 |
| LeakyReLU-24 | [8, 256, 4, 4] | 0 |

Table 1(b): Generator Layers

| Layer | Output Size | Parameters |
| --- | --- | --- |
| Linear-1 | [8, 16384] | 16,54,784 |
| ConvTranspose2d-2 | [8, 512, 8, 8] | 83,89,120 |
| BatchNorm2d-3 | [8, 512, 8, 8] | 1,024 |
| ReLU-4 | [8, 512, 8, 8] | 0 |
| ConvTranspose2d-5 | [8, 256, 16, 16] | 20,97,408 |
| BatchNorm2d-6 | [8, 256, 16, 16] | 512 |
| ReLU-7 | [8, 256, 16, 16] | 0 |
| ConvTranspose2d-8 | [8, 128, 32, 32] | 5,24,416 |
| BatchNorm2d-9 | [8, 128, 32, 32] | 256 |
| ReLU-10 | [8, 128, 32, 32] | 0 |
| ConvTranspose2d-11 | [8, 64, 64, 64] | 1,31,136 |
| BatchNorm2d-12 | [8, 64, 64, 64] | 128 |
| ReLU-13 | [8, 64, 64, 64] | 0 |
| ConvTranspose2d-14 | [8, 32, 128, 128] | 32,800 |
| BatchNorm2d-15 | [8, 32, 128, 128] | 64 |
| ReLU-16 | [8, 32, 128, 128] | 0 |
| ConvTranspose2d-17 | [8, 3, 256, 256] | 1,539 |

## 4. Objective Function

The mDCGAN is trained using the Binary cross entropy loss function. The ground true labels are real or fake, and the input is a probability obtained from the sigmoid function. Equation (2) describes the expression used to compute the loss value,

$$L = -\frac{\sum_{n=1}^{n=N}[y_n \cdot \log(z_n) + (1 - y_n)\log(1 - z_n)]}{N} \quad (2)$$

In equation (2), $y_n$ represents the ground truth label for real or fake, $N$ is the number of samples, and $zn$ is the output of the sigmoid activation. The loss function mentioned in equation (2) is the negative log likelihood function. Both the generator and discriminator are trained using this loss function. In order to set up an adversarial training loop, the loss of the generator and discriminator are modelled as per the min max algorithm[8]. Equation (3) and (4) describe the generator and discriminator loss as described in [40], where $L_D$, $L_G$ represent loss of the discriminator and generator respectively.

$$L_D = \frac{1}{N}\Sigma_{n=1}^{n=N}[\log(D(x^{(n)})) + \log(1 - D(G(z^n)))] \quad (3)$$

$$L_G = \frac{\Sigma_{n=1}^{n=N} \log(1 - D(G(z^n)))}{N} \quad (4)$$

While training the discriminator, we tried concatenating the real and fake images with their respective labels before passing it into the discriminator. However, this approach resulted in unstable training where the discriminator loss experienced a steep drop to a value around 0 within the first 5 to 10 epochs as shown in figure 3. To tackle this issue, the real and fake images were passed separately to compute their loss and their arithmetic mean was computed. This approach provided us with a much more stable training loop which we describe in section 4. Equation (5) describes our approach for training the generator.

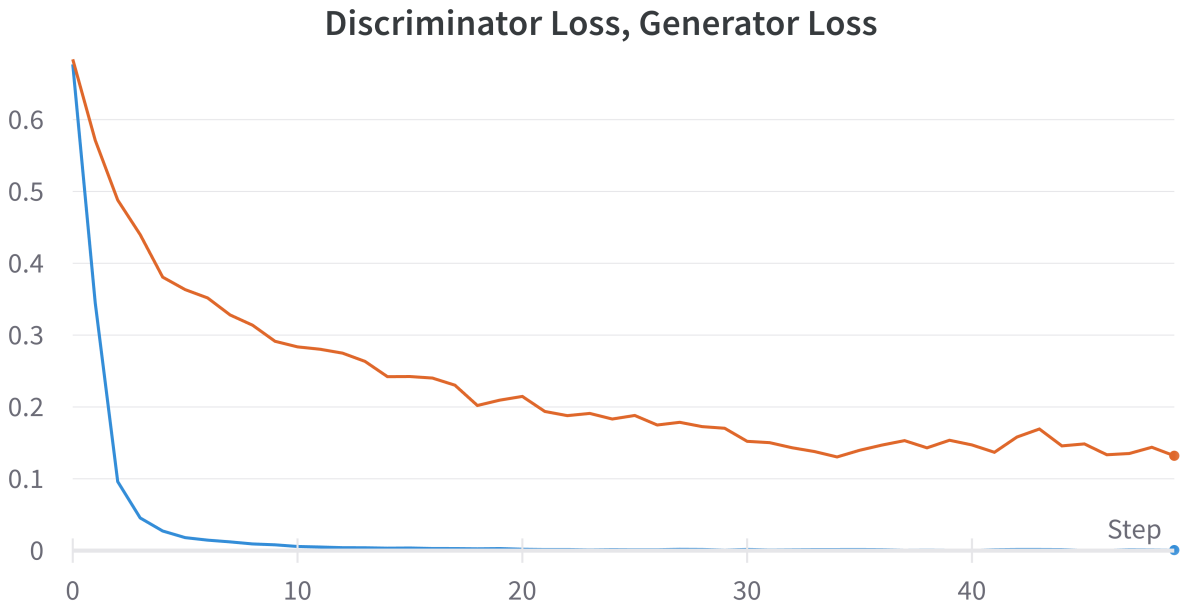

Figure 3: Unstable Training using combined distribution of Real and Fake Samples

$$Loss_D = \frac{Loss_{DR} + Loss_{DF}}{2} \quad (5)$$

where, $Loss_{DR}$, $Loss_{DF}$ represents the discriminator loss on passing real and fake samples respectively.

# 5. Optimisation Algorithm

Both the generator and discriminator were trained using the Adam optimiser[41]. Adams optimiser uses Momentum and the exponential moving average of gradients from RMSProp in order to attain convergence quickly. The following parameters of Adam's optimiser were used for our study:

i. $\beta_1$, $\beta_2$ which are the exponential decay rates of the first and second moment of the gradients respectively were set to 0.5 and 0.999.

ii. The learning rate parameter was set to 0.002

iii. $\epsilon$ set to its default value of $10^{-8}$

# IV. Experiment and Results

## 1. Experiment Setup

The experiments were performed on a 2022 model Macbook Pro, with Apple Silicon M2 chip. The configurations of the system are inclusive of an 8GB unified memory with 256GB Hard disk storage. The system has 8 Core CPU, which has a split-up of 4 performance cores and 4 efficiency cores, and a 10 core GPU. In addition to this, the system has a 16-core neural engine.

The models used for the experiment were trained on PyTorch 2.0.1 with the help of helper image transforms package Torchvision version 0.15.2. The Beta version of Torchvision was used for image transformations. PyTorch makes use of the Metal Performance Shaders backend for accelerated GPU training. This extends the PyTorch framework, providing scripts required to run operations on a Mac. In addition, libraries such as Numpy, Matplotlib and Scikit-learn were also used to conduct the experiments During experimentation the train and test metrics on Weights and Biases were continuously recorded which provided insightful plots of train and test curves.

In order to prevent the MPS backend from running out of memory and memory leaks, the following steps were put to practice in the experiment process:
- Batch size restricted to 32
- Usage of del[42] to delete tensors after a train and test step
- Clear cache blocks of the MPS backend using mps.empty_cache()[42]

## 2. Dataset

The dataset chosen for experimentation is the Abstract Art Gallery obtained from Kaggle[43]. It is a diverse dataset comprising of 2782 images of different paintings, each with a distinct colour scheme and pattern scraped from various web sources. From initial visualisation most images have used strokes of green, red, black, and orange. For the purpose of training, randomly sampled 2000 images were used from the dataset while maintaining the overall distribution of the images. The images have varied sizes and resolutions such as 1024X2048, 1024X1024, 512X512 etc, for the purpose of uniformity we resized all images to

1024X1024. All images were smoothened using standard filters such as Gaussian and Median, with no data augmentation.

# 3. Environment

Tuning of hyper-parameters especially for training Generative Adversarial Networks play a crucial role in recognising patterns and generating stable images. GANs are highly susceptible to unstable training, modal collapse and noise, therefore it is important to fine tune hyper-parameters optimally[44]

Table 2: Summary of Training Environment Parameters

| Sr. No | Hyperparameter | Value |
| --- | --- | --- |
| 1 | Number of Epochs | 1000 |
| 2 | Adam's Beta 1 | 0.5 |
| 3 | Adam's Beta 2 | 0.999 |
| 4 | Learning Rate | 0.002 |
| 5 | Batch size | 32 |

Table 2 describes the parameters used to obtain stable training of mDCGAN. The Adams Beta 1 and Learning Rate parameter were tuned according to the parameters mentioned in [38]. Batch size was set to 32 taking memory constraints of the system environment into account.

# 4. Training Analysis

The model was trained according to the training parameters mentioned in section 2.3. The work involved training the two modules of mDCGAN together: the generator and the discriminator. Figure 3 describes the training curves of the generator and discriminator. Blue represents generator loss and orange represents discriminator loss. From Figure 3, it can be seen that a close to ideal situation of GAN training is achieved in the training of mDCGAN. The generator and discriminator both try to gain the upper hand against each other. The learning function both the generator and discriminator are similar upto epoch 300 after which they diverge. The loss values oscillate about the value of 0.015 to 0.016 for both the discriminator and generator. No upper or lower bounds on loss were observed due to the unstable training conditions of GANs. It is to note that the loss function is not a metric for training but only used for plot visualisation purposes. After epoch 500 we get noisy outputs due to overfitting which is further discussed in the next section.

# 5. Analysis of Generated Brush Stroke Patterns

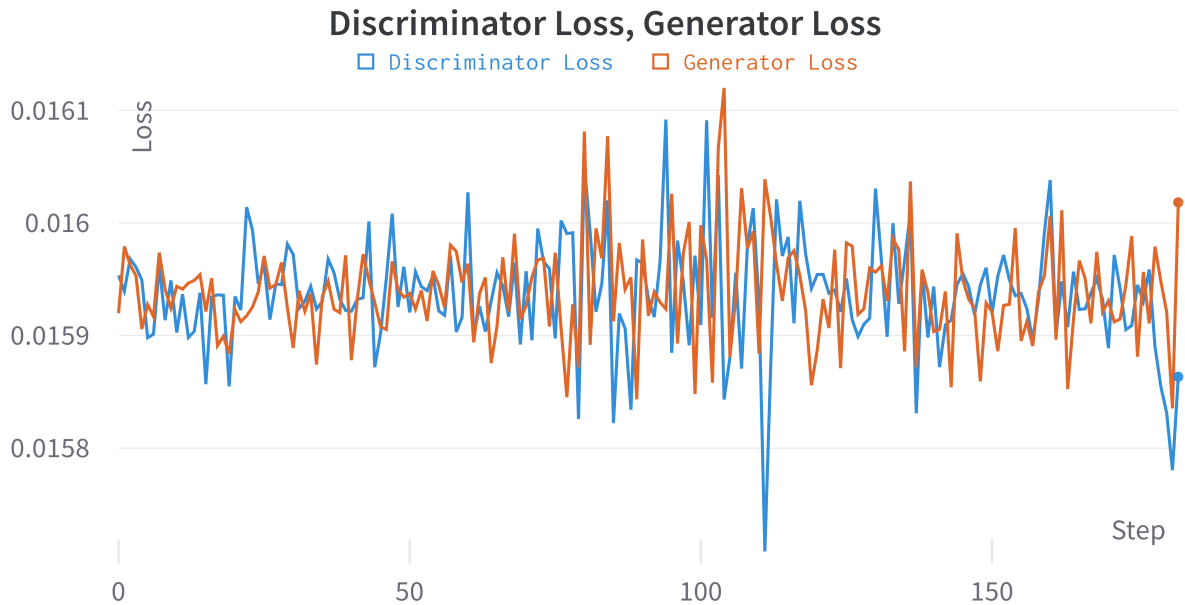

Figure 4: Generator and Discriminator Loss

The outputs obtained after generating images are as shown in Figure 5(a) to Figure 5(e). As noted in section III, the input is a random noise vector generated from a Uniform distribution of 100 dimensions. It can be observed that the model generates multiple different brush stroke patterns using the colours present in the original dataset. We observe the dominance of black strokes, which is a prevalent feature in modern day abstract art[45]. Other patterns include brush strokes of lighter and darker shades of blue, and tin shades of green and red. These colours usually dominate in abstract art paintings symbolising various emotions and depictions of artists. The generated brush strokes have unique features with respect to gradients, edges and directional patterns. It can be observed that when red is the dominant colour, blue is usually used as a complementary colour along with strokes of purple and black. However in certain paintings, dark blue is observed as the dominant colour which is usually complemented by a black shade to make the painting a blue-purple combination.

# 6. Space Exploration through Random Walks

This section describes the experiment of exploring the latent space. The distribution of the patterns lies in the latent space, which is a multi-dimensional abstract space that encodes the information of the outside world. The latent space of the brush stroke and colour patterns generated by mDCGAN is explored by performing algebraic vector operations to discover new patterns and colours. Figure 4.3 depicts the argument mentioned above. $V_1$, $V_2$, $V_3$ are the initial vectors generated by Generator of mDCGAN. Later, a set of vector operations are performed to find new vectors in the art space. The relationships between the image vectors are modelled as per the parallelogram law of vector addition. The following equations (6), (7) and (8) describe the operations performed.

$$V = V_2 + V_3 - V_1 \quad (6)$$
$$V = V_1 + V_2 - V_3 \quad (7)$$
$$V = V_1 - V_2 + V_3 \quad (8)$$

On performing the above set of operations, the rise of new colour shades and patterns, and a mathematical relation between different shades of colours is observed. The development of colour shades such ad blue, yellow and lighter shades of green is seen through this process. Hence it is possible to develop vector relationships between colour contrasts, analogous to logical relationships between words captured by word embeddings[46]. From figure 6(a) and 6(b), we can derive the following colour-based relationships described by equations (9) and (10)

$$Green = 2 * Red - Black \quad (9)$$
$$Blue_{light} = Green + Blue_{dark} - Black \quad (10)$$

It is to note that the above mentioned equations are derived from a pure qualitative basis. The Experiment performed in figure 6(b) was performed mid-training i.e after 175 epochs. The results of experiment 6(b) show the presence of light, dark blue and green shades at early stages of training. Figure 6(c) show the results of our third random walk experiment. We can derive the relation mentioned in (11),

$$Yellow = Red_{dark} + Red_{Light} - Black \quad (11)$$

## 7. Unstable GAN Outputs with Time

### 7.i Qualitative and Numerical Analysis

The model was trained upto 500 epochs and began result visualisation after the $150_{th}$ epoch. It was noted that after epoch 250, the patterns were distorted with highly pixelated outputs resulting in heavy noise and feature loss. This is probably due to the weights of the generator trying to overcorrect resulting in large updates when the discriminator begins to outperform the generator. Figure 7(a) and 7(b) highlight this observation further.

We notice that the black brush strokes are still clear while the other colours such as blue, green, yellow etc try to overfit the canvas resulting in high distortion A quantitative analysis was performed on the noisy image distribution by comparing it with the stable distribution. Table 3 contains the metrics used to support our qualitative argument. **It is to note that for the SNR computation, the experiment considers noise as output of generator after epoch 275 and stable images at epoch 250 as signal.**

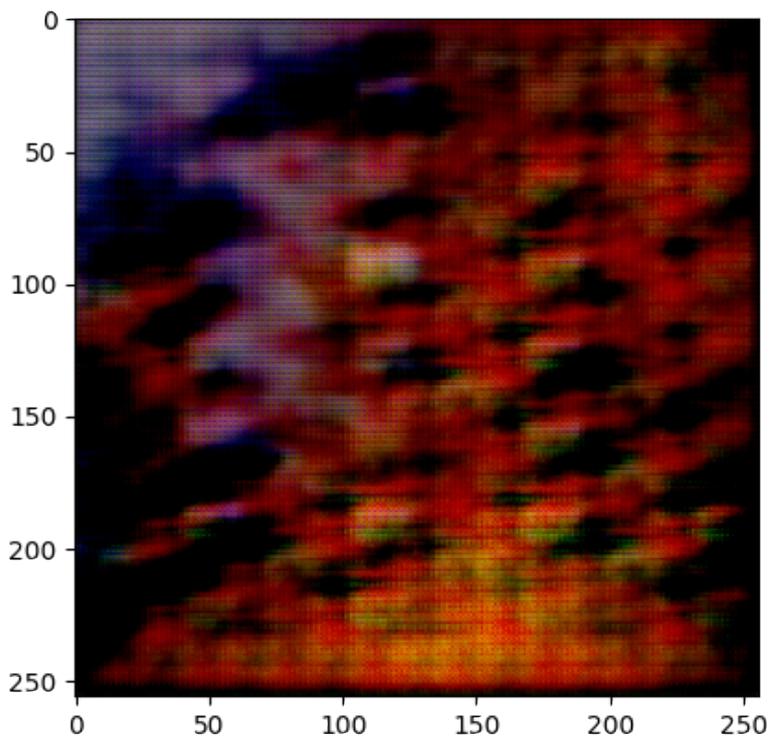
Figure 5(a)

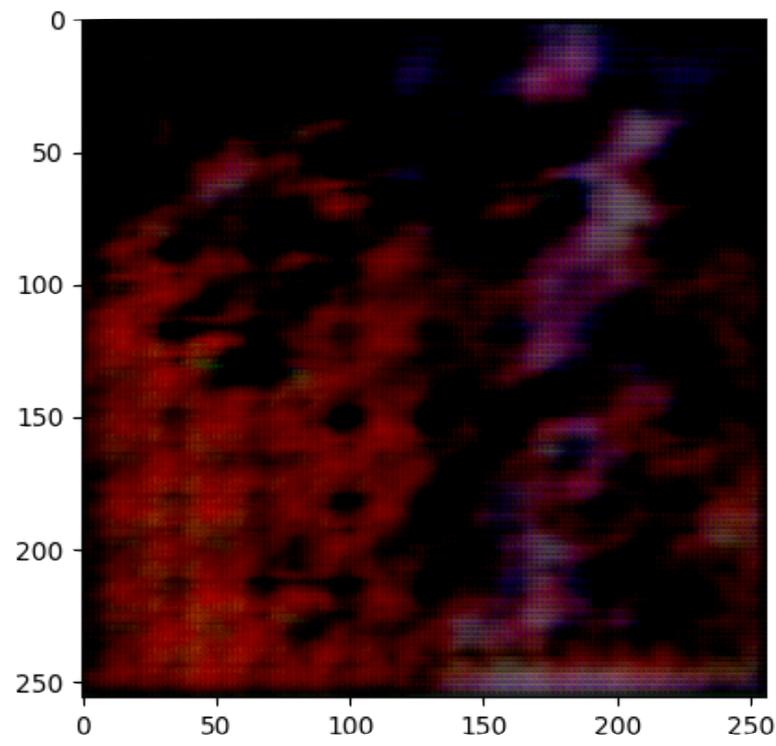
Figure 5(b)

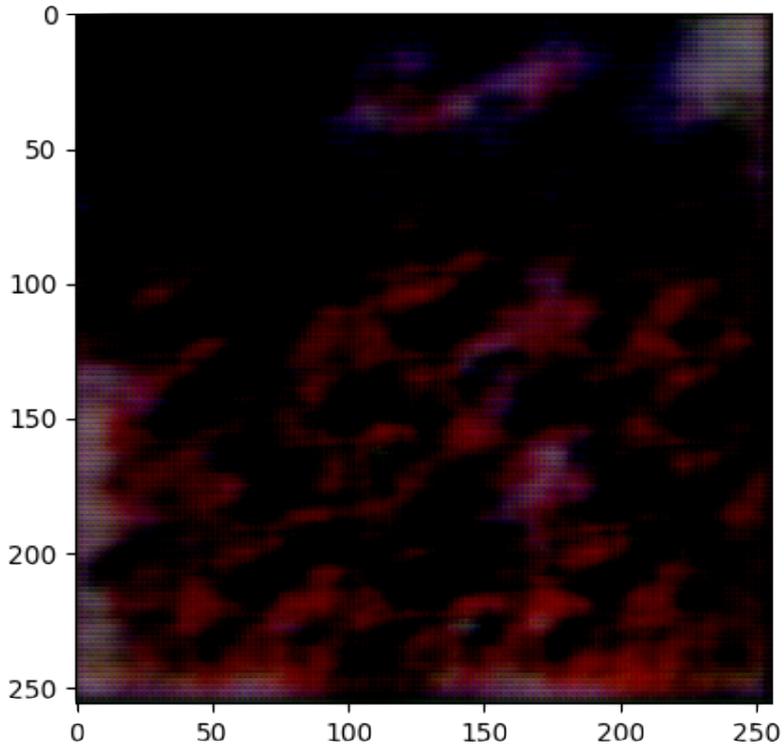
Figure 5(c)

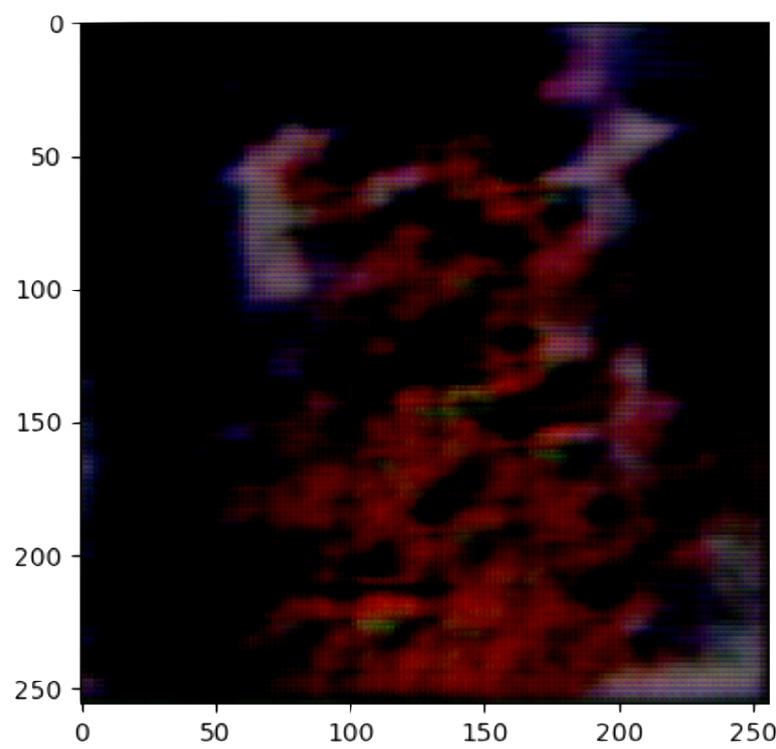
Figure 5(d)

Table 3: Comparison between Stable and Erroneous Brush Stroke Colour Distribution

| Metric | Equation | Value |
| --- | --- | --- |
| Signal to Noise Ratio | $SNR = 10 \log_{10} \dfrac{power_{signal}}{power_{noise}}$ (12) | 7.081 |

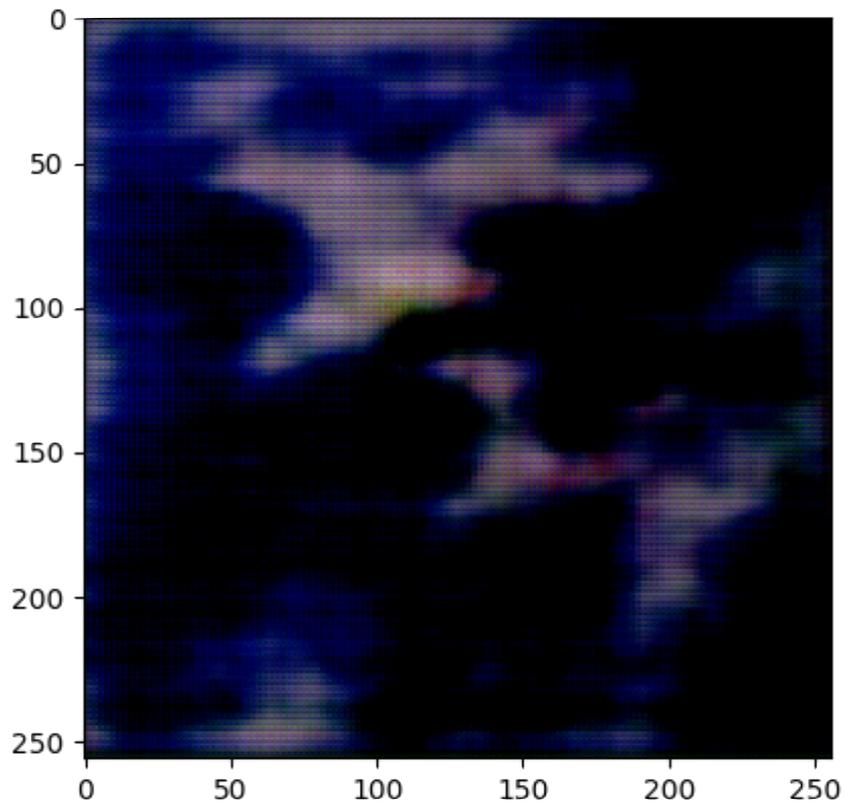

Figure 5(e)

| Metric | Equation | Value |
|---|---|---|
| $L_2$ Distance | $d(x, y) = \sqrt{\Sigma_1^n (x_i - y_i)^2}$  (13) | 35.2102 |
| $L_1$ Distance | $d(x, y) = \Sigma_1^n |x_i - y_i|$  (14) | 28.6706 |

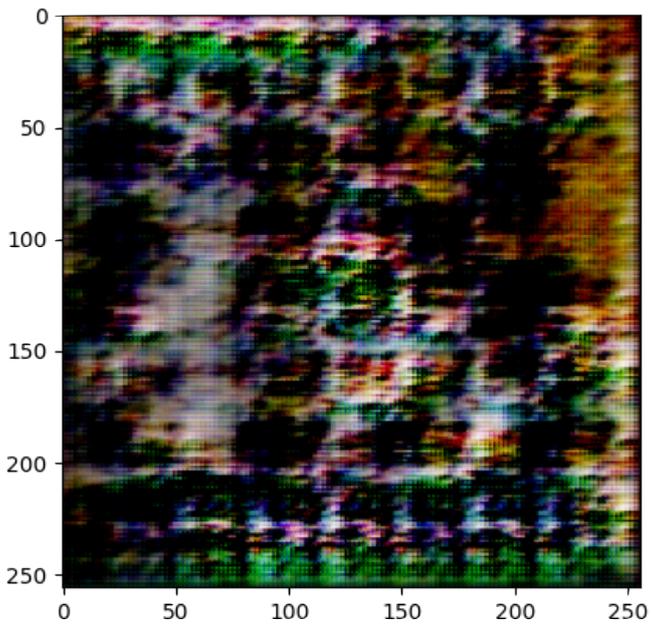

Figure 7(a)

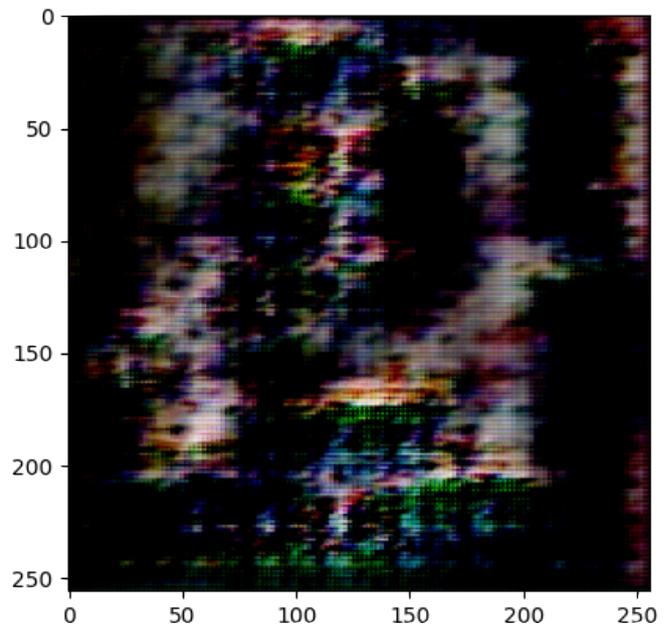

Figure 7(b)

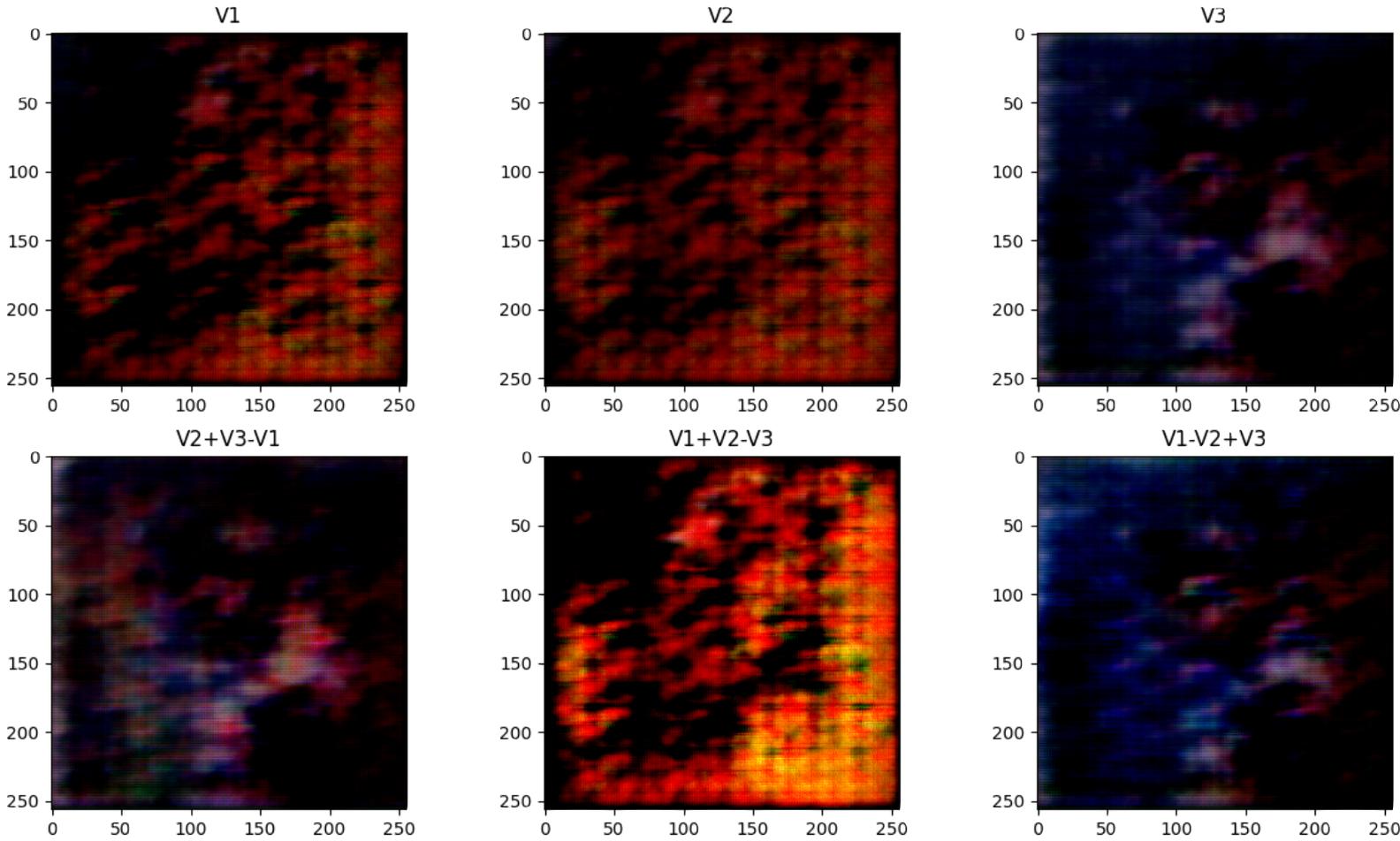

Figure 6(a): Random Walk Experiment

Signal to Noise ratio greater than 0 represents a strong signal power. The SNR proves the argument that the images after epoch 250 are highly distorted as noted qualitatively. In addition, $L_2$ and $L_1$ distances were computed between the two distributions. The values mentioned in Table 3 for the above two metrics show that the two distributions are dissimilar thereby proving the qualitative argument. It is to note that these are informal metrics to measure image quality and GAN performance.

## 7.ii Hypothesis Test of Difference in Variance using F-Test

To formalise the argument that there is a significant difference between the art space distribution before and after epoch 250, an F-Test is conducted to compare the variance of the two distributions. The test statistic for the samples i.e sample variance is computed for around **101 samples**. Table 4 contains the sample mean and variance for both the distributions.

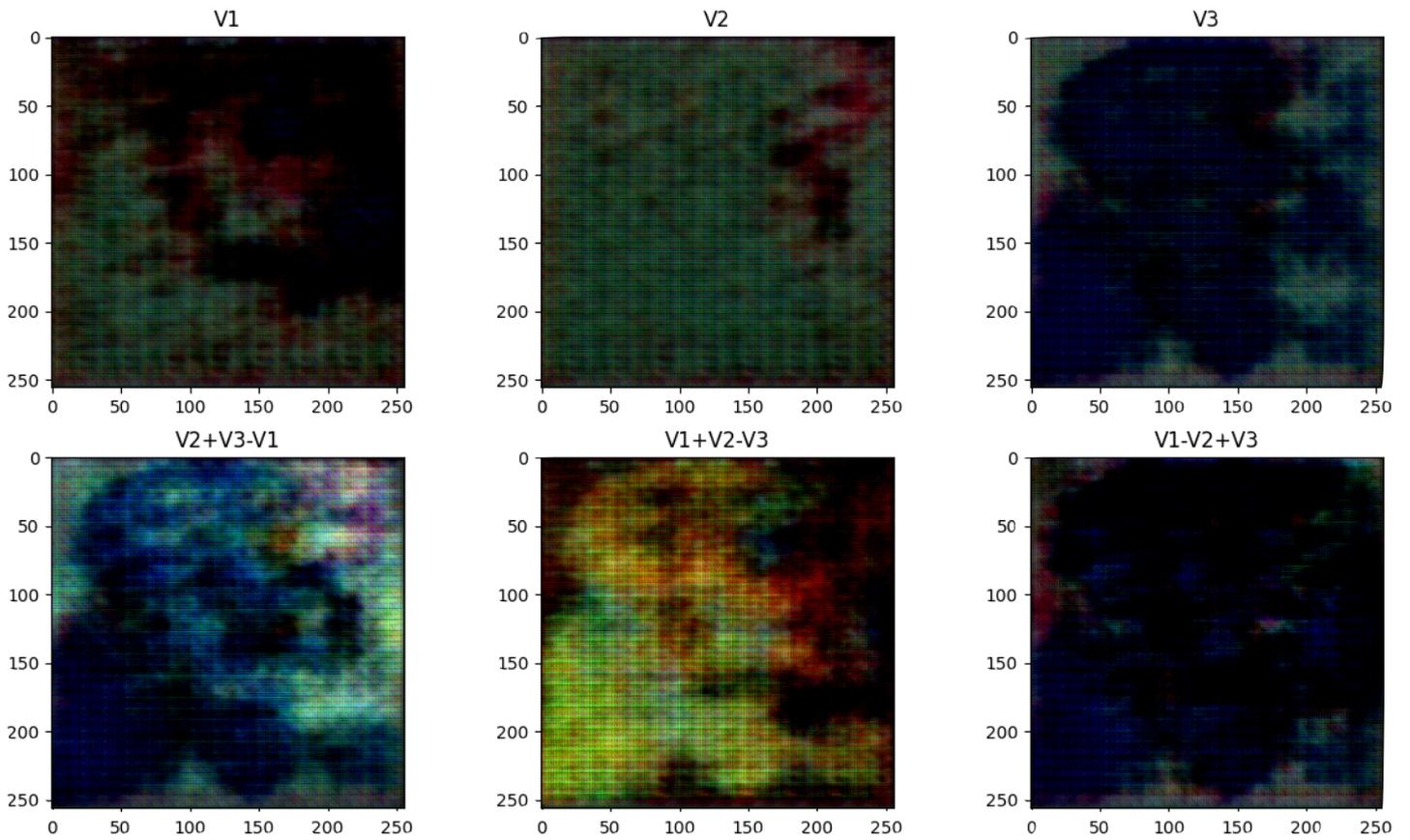

Figure 6(b): Random Walk Experiment: Mid Training

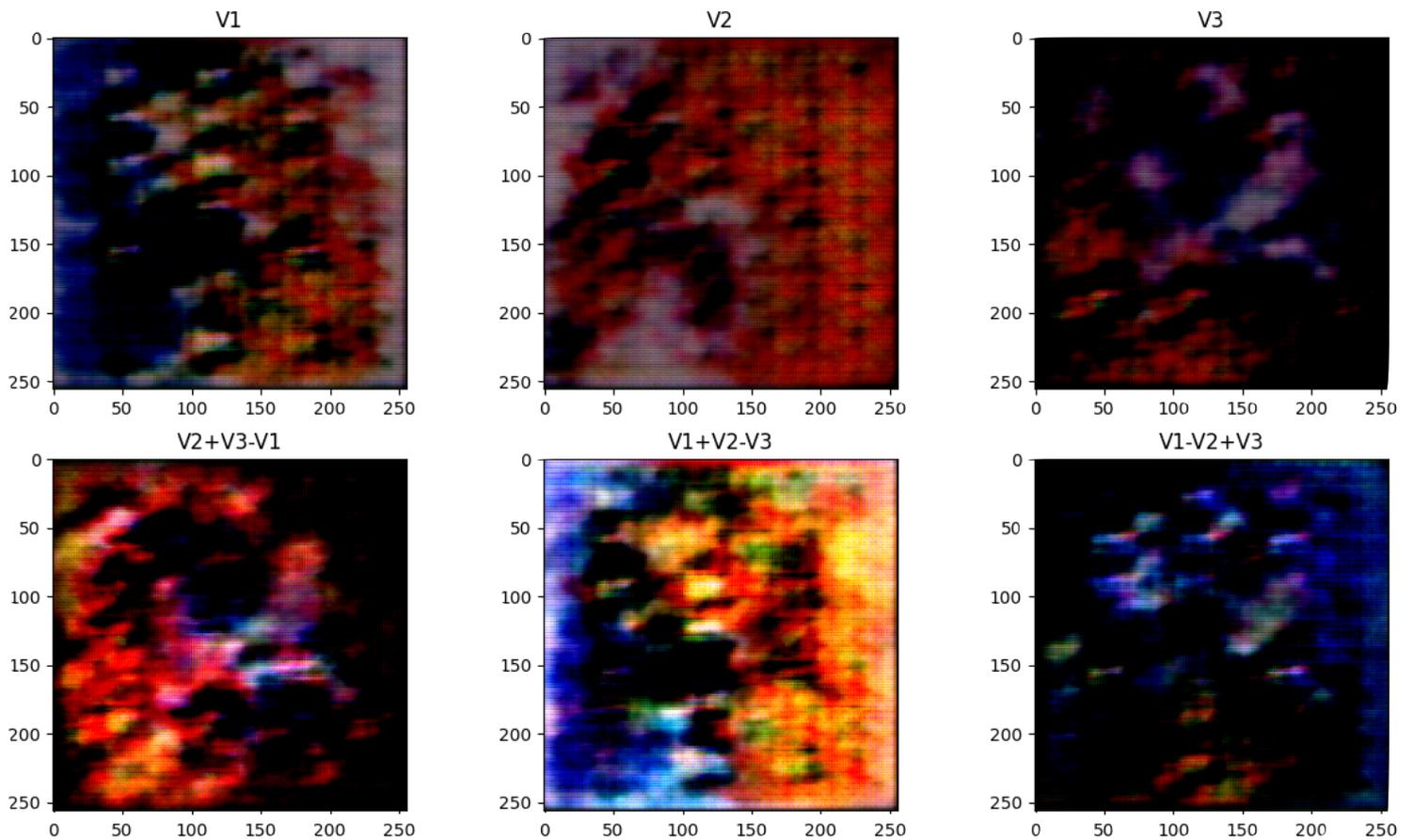

Figure 6(c): Random Walk Experiment

Table 4: Sample mean and standard deviation

| $n_1 = n_2 = 101$ | Generated Samples after Epoch 250(Sample 1) | Generated Samples at Epoch 250(Sample 2) |
|---|---|---|
| **Sample Mean** | 123.68409 | 125.07779 |
| **Sample Standard Deviation** | 40.81453 | 31.979033 |

The Hypothesis is defined as follows let $\sigma_1, \sigma_2$ be the standard deviations of the two distributions, $H_0$ and $H_A$ represent the null and alternative hypothesis respectively. We define the test as follows:

$$H_0 : \sigma_1^2 = \sigma_2^2$$
$$H_A : \sigma_1^2 \neq \sigma_2^2$$

The Null Hypothesis states that there is no significant difference between the standard deviations of the two samples, while the Alternative hypothesis states otherwise. The test statistic $F$ is computed using equation (15).

$$F = \frac{\sigma_1^2}{\sigma_2^2} \quad (15)$$

$\sigma_1, \sigma_2$ represent the sample standard deviation for samples 1 and 2 respectively. On inserting the values listed in Table 4 in equation (15), the value of test statistics $F$ is 1.629. Confidence intervals of 95% and 99% are considered, therefore the level of significance $\alpha$ is 0.05 and 0.01 respectively. The obtained results on computing the test statistic along with the critical value of the confidence intervals are listed in Table 5.

Table 5: Critical Value and Test Statistic

| degrees of freedom($n - 1$)=100 | Level of Significance $\alpha$ | Critical Value $c$ | Test Statistic $F$ |
|---|---|---|---|
| 95% Confidence | 0.05 | 1.392 | 1.629 |
| 99% Confidence | 0.01 | 1.598 | 1.629 |

For a two tailed test, the rejection region is defined as $F \leq -c \text{ or } F \geq c$ where $c$ is the critical value. For both the intervals, $F > c$ therefore null hypothesis $H_0$ is rejected. Hence the qualitative argument that there is a significant difference between the variances of the two distributions is proved.

# V. Conclusion and Further Work

In our study, we generate colour and brush stroke patterns in abstract art by using a modified version of a DCGAN to fit our needs. We see that darker colours such as black, dark red, purple and dark blue are very popular among abstract artists. We observe the use of dark brush strokes complemented by lighter colours in the palette. Later we performed a random walk to explore the latent space comprising of multicoloured strokes and were, qualitatively, able to extract vector relationships between colours through multiple random walk experiments performed at the end of training and at its early stages. Furthermore we analyse unstable distorted GAN outputs after epoch 250 in training by a statistical analysis using Signal to Noise Ratio and distance between distributions metrics and hypothesis testing of difference in variance to show that there is a significant difference between sample distributions before and after epoch 250. Further work can be carried out by employing larger GANs such as StyleGANs for higher resolution outputs while techniques such as edge detection, gradient-based study of brush strokes etc can explored apart from latent space exploration.